\documentclass[11pt]{article}

\usepackage[preprint]{acl}
\usepackage{times}
\usepackage{latexsym}
\usepackage[T1]{fontenc}
\usepackage[utf8]{inputenc}
\usepackage{microtype}
\usepackage{inconsolata}
\usepackage{graphicx}

\usepackage{booktabs}
\usepackage{multirow}
\usepackage[table]{xcolor}
\definecolor{imprcolor}{HTML}{EFEBF7}
\definecolor{lightgraycolor}{HTML}{E3E3E3}

\usepackage{amsmath}
\usepackage{amssymb}
\usepackage{mathtools}
\usepackage{amsthm}

\usepackage[capitalize,noabbrev]{cleveref}
\usepackage[table, svgnames]{xcolor} 
\usepackage{xspace}
\usepackage{enumitem}
\usepackage{amsmath}
\usepackage{amssymb}
\usepackage{float}

\newcommand{\red}[1]{\textcolor{red}{#1}}

\newcommand{\name}{\texttt{T2SP}\xspace}
\newcommand{\define}[1]{\vspace{0mm}\noindent{{\textbf{#1.}}}}
\newcommand{\ie}{\emph{i.e.,}\xspace}
\newcommand{\eg}{\emph{e.g.,}\xspace}

\usepackage{caption}
\usepackage{xcolor}
\usepackage{listings}
\usepackage[most]{tcolorbox}

\lstdefinestyle{tsprogram}{
    basicstyle=\ttfamily\scriptsize, 
    keywordstyle=\bfseries\color{blue!60!black},
    stringstyle=\color{green!40!black},
    commentstyle=\itshape\color{gray},
    numbers=none,
    breaklines=true,
    columns=fullflexible,
    keepspaces=true,
    showstringspaces=false,
    frame=none,
}

\newtcblisting{tsprogrambox}[1][]{
    listing only,
    colback=gray!4,
    colframe=gray!35,
    arc=1mm,
    boxrule=0.35pt,
    left=0.5mm,
    right=0.5mm,
    top=0.5mm,
    bottom=0.5mm,
    width=0.9\linewidth,
    center,
    fonttitle=\scriptsize\bfseries, 
    coltitle=black,
    listing options={
        basicstyle=\ttfamily\scriptsize,
        breaklines=true,
        columns=fullflexible,
        keepspaces=true,
        showstringspaces=false
    },
    #1
}

\usepackage{listings}
\usepackage{xcolor}

\lstdefinestyle{promptstyle}{
    backgroundcolor=\color{gray!5},
    frame=single,
    rulecolor=\color{gray!50},
    basicstyle=\tiny\ttfamily,
    breaklines=true,
    breakatwhitespace=true,
    columns=fullflexible,
    numbers=none,
    keepspaces=true,
}

\title{Representing Time Series as Structured Programs for LLM Reasoning}

\author{
 \textbf{Jaeho Kim\textsuperscript{1}\thanks{Equal Contribution.}},
 \textbf{Changjun Oh\textsuperscript{1}\footnotemark[1]},
 \textbf{Seokhyun Lee\textsuperscript{1}},
  \textbf{Irina Rish\textsuperscript{2}},
 \textbf{Changhee Lee\textsuperscript{1}\thanks{Corresponding author.}}
\\
 \textsuperscript{1}Korea University, 
  \textsuperscript{2}Mila, University of Montreal 
\\
}

\begin{document}
\maketitle
\begin{abstract}
Large language models (LLMs) have demonstrated strong reasoning and instruction-following capabilities, making them potentially powerful tools for time-series analysis. However, time series lie outside their native textual modality, raising a fundamental question: \textit{how should time series be represented so that LLMs can reason about them effectively?} Existing work typically serializes raw numerical sequences or fine-tunes pre-trained LLMs on time-series data. These approaches place the burden of extracting temporal structure directly on the LLM, creating a modality mismatch that often degrades performance on long sequences and introduces substantial computational overhead. In this work, we introduce Time-Series-to-Structured-Program representation~(\name), a deterministic, training-free method that represents a time series as a \emph{structured symbolic program}. \name decomposes time series into trends, periods, and salient events, expressing them in a \emph{program-friendly} format aligned with the textual and code-like modalities on which LLMs are natively trained. By shifting temporal-structure extraction from the model to the representation itself, \name enables off-the-shelf LLMs to leverage their existing reasoning capabilities for time-series understanding. We evaluate \name on three reasoning tasks -- editing, captioning, and question answering -- where it consistently improves performance, reduces reasoning time, and lowers failure rates compared with raw-string representations. Our results demonstrate that \name provides an effective interface between time series and LLMs.

\end{abstract}

\section{Introduction}
\label{sec:intro}
\begin{figure}[t]
\begin{center}
\hspace*{-0.7cm}\includegraphics[width=1.2\columnwidth]{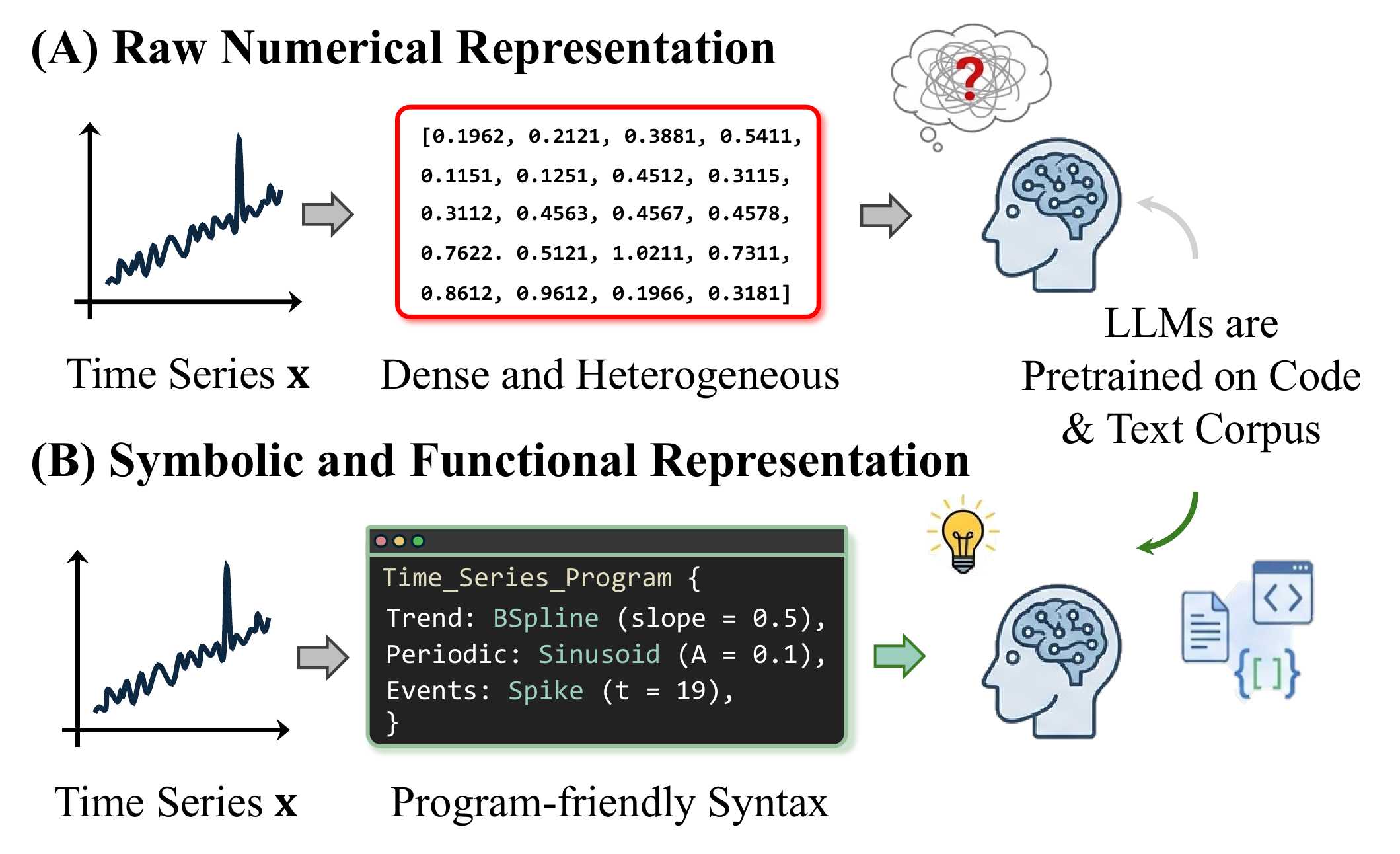}
\vspace{-20pt}
    \caption{\define{Representations matter for LLM-based time-series reasoning} (A) Raw numerical sequences are dense and heterogeneous, forcing LLMs to infer the temporal structure value by value. (B) LLMs are pretrained on a vast corpus of natural language texts and code representations~\cite{touvron2023llama}, making them well-suited for reasoning over symbolic and functional forms~\cite{gao2023pal, chenprogram}.}
    \label{fig:motivation}
    \vspace{-20pt}
\end{center}
\end{figure}

Large language models (LLMs) have achieved strong success across diverse domains such as mathematics~\cite{hendrycks2measuring}, coding~\cite{chen2021evaluating}, and reasoning~\cite{lu2024ai}, with their ability to follow instructions and reason over textual information increasingly approaching human-level performance~\cite{phan2025humanity}. Building on these successes, a growing line of work has begun to apply LLM capabilities to time-series analysis~\cite{jin2024position}, asking LLMs to interpret, edit, and reason about temporal data, beyond classical classification and forecasting. However, LLMs that excel elsewhere have been shown to struggle with extracting meaningful structures from raw time series~\cite{merrill2024language}, and consequently fail on seemingly simple time-series analysis tasks. We argue that this gap arises not from a deficiency of the model, but from a representation mismatch~(shown in \cref{fig:motivation}).

Yet the dominant response in the literature has focused on the \emph{model side}  of the interface, taking three mainstream directions. The first and most direct method is to feed the \emph{raw time series as a string}, requiring the LLM to interpret raw values directly. Although this preserves the full fidelity of the sequence, this approach breaks down as sequence length grows~\cite{fons2024evaluating}, and LLMs are known to struggle with raw numerical inputs~\cite{zhou2025fone}. A second line of work converts time series into visual plots and uses vision-language models~(VLMs) for analysis. While vision-based models are effective at capturing the holistic shape of a series in a form readable to humans, they struggle to capture structural components such as periodicity or localized events, limiting their applicability to tasks that require a fine-grained understanding~\cite{sen2025bedtime}. A third approach is to fully or partially fine-tune the LLM on time-series or domain-specific representations~\cite{wang2025chattime}. Unfortunately, such methods are computationally expensive and not applicable to the strongest closed-source LLMs accessible only via APIs, limiting their scalability. These approaches share a commonality: \emph{they push the burden of time-series understanding onto the model} -- asking the model to read numerical sequences, switch modalities, or even retrain the model. We take the opposite view. \textit{We argue that the model is already highly capable}; what is needed is a representation that (i) preserves the informative structure of time series with sufficient fidelity, and (ii) exposes that structure in a symbolic form the LLM can natively parse and reason over. 

How, then, \emph{should time series be represented to satisfy both criteria?} Our answer is that time series should be re-expressed in a \emph{program-friendly} and \emph{structured} format that LLMs can already reason over effectively. Whereas a list of raw numbers forces the model to understand the temporal structure value by value, a structural representation of time series aligns with the symbolic and code-like expressions LLMs handle fluently~\cite{gao2023pal, achiam2023gpt}. Motivated by this perspective, we introduce the Time-Series-to-Structured-Program representation~(\name), a novel time-series representation method for interfacing time series with LLMs. \name \emph{deterministically} decomposes a raw series into components of trend, periodic, event, and residual components, and expresses them as a structured program using program-like syntax that is readable by both humans and LLMs. Our contributions are as follows:
\begin{itemize}[leftmargin=13pt, itemsep=-0.5pt]
    \vspace{-5pt}

    \item We propose \name, a time-series representation with a structured and program-like syntax that aligns with the modalities on which LLMs were natively trained, enabling them to reason directly over temporal structure rather than inferring it from raw numerical sequences.

    \item \name is deterministic, invertible, training-free, and compatible with off-the-shelf LLMs, including API-only models. Building on \name, we perform time-series editing, captioning, and question answering through a unified interface -- without tool calls, agentic loops, or fine-tuning.
    
    \item Across time-series tasks and LLM models, \name improves reasoning performance, reduces inference time, and lowers LLM response failure rates, compared with raw-string representations, while scaling gracefully with sequence length.
\end{itemize}

\section{Related Works}
\subsection{Time-series Representations for LLMs}
\textbf{Adapting the model to time series} has been a mainstream research direction, aligning with the standard deep learning paradigm in which data is often treated as fixed. Within this direction, two threads have emerged.

The first line of work introduces trainable parameters into the time-series-to-LLM pipeline, whether through training encoders and adapters, or even updating the LLM itself. For instance, TEST~\cite{sun2024test} introduces a learnable encoder and prompt embedding method to enable the frozen LLM to accept time-series input. ChatTime~\cite{wang2025chattime} introduces an expanded vocabulary set to process time series and fine-tunes the embedding layers of LLMs. Since updating these parameters risks degrading the LLM's general reasoning capability, ChatTime requires an additional pre-training and fine-tuning procedure to stabilize performance. Time-MQA~\cite{kong2025time} and ARTIST~\cite{messica2026adaptive} both perform supervised fine-tuning on an open-source LLM such as Llama-3 8B and Qwen3-4B for reasoning tasks. While these methods offer a dedicated methodological contribution for time-series tasks, they require parameter updates and a non-trivial training strategy. Moreover, since supervised fine-tuning~(SFT) is typically feasible only on small open-source models, it remains incompatible with the strongest closed-source LLMs, limiting its scalability.  

The second line of work adapts the model by switching to an 
entirely different modality: time series are rendered as visual plots~\cite{ni2025harnessing} and processed by vision-language models~(VLMs). For instance, Time-VLM~\cite{zhong2025time} renders time series as plots and uses a VLM encoder to extract embeddings. However, recent works~\cite{wang2025chattime, sen2025bedtime} verify that performance degrades monotonically with image resolution, and that VLM-based approaches struggle with fine-grained numerical tasks~\cite{ding2026llatisa}.

\noindent\textbf{Adapting the representation} takes the opposite stance: the LLM is fixed, and time series are instead paraphrased in a form that the model can understand. This makes the approach training-free and directly applicable to closed-source LLMs. The most direct instantiation is to provide time series as raw strings, and a number of works~\cite{kong2025time, xie2025chatts, sen2025bedtime, messica2026adaptive} adopt such string-based representations as the primary interface between time series and LLMs, owing to their simplicity and intuitiveness. For instance, LLMTIME~\cite{gruver2023large} feeds time series into LLMs as raw strings with specialized preprocessing steps (\eg per-digit tokenization), and PromptCAST~\cite{xue2023promptcast} rephrases the series into a structured prompt that is directly used by the LLM. However, string-based representations also have inherent limitations: tokenization cost scales with both sequence length and numerical precision, which degrades performance on long sequences~\cite{sen2025bedtime}. Nevertheless, the broad applicability of string representation across LLMs and the ease of use make it the most natural baseline for evaluating any time-series interface. Since \name operates on the same string-level and requires no additional training, we adopt string-representation as our primary comparison and demonstrate the benefits of \name across a diverse set of LLMs.

\subsection{Time-series Reasoning}
Time-series reasoning is the task of utilizing the reasoning capabilities of LLMs with their contextual understanding to ``reason'' about time series through language. Recently, a number of time-series reasoning tasks~\cite{kong2025time, wang2025chattime, sen2025bedtime, qiu2026instruction} have been proposed, and at the core of these tasks lies the ability to grasp the underlying structure of a series rather than its raw values. To this end, we focus on three representative tasks that most directly demand such structural understanding: time-series editing, captioning, and question answering. Specifically, time-series editing~\cite{jing2024towards, qiu2026instruction}  focuses on editing a source time series into a target series that encompasses user-specified attributes. This task is primarily used to examine counterfactual, or \textit{what-if} scenarios. Time series captioning, also referred to as open generation~\cite{sen2025bedtime}, is the task of describing the time series with text. Enabling a precise description of the time series plays an important role in multi-modal learning. Lastly, question answering~\cite{wang2025chattime} requires the model to answer natural-language questions about a given time series, verifying its ability to comprehend and reason over temporal data. In this work, we evaluate \name on all three tasks, demonstrating that explicit structural representation consistently benefits time-series reasoning across diverse LLMs.
\begin{figure*}[t]
\begin{center}\center{\includegraphics[width=0.95\textwidth]{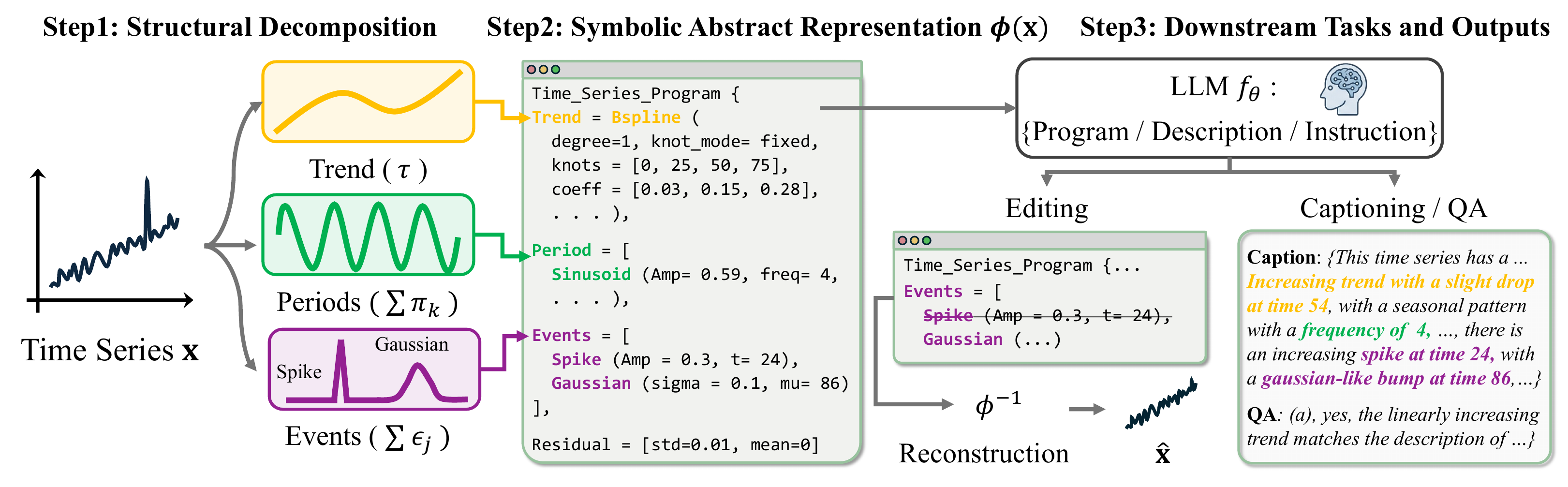}}
\vspace{-10pt}
\caption{\textbf{Overview of \name.} A time series is decomposed into structured components -- trend, periods, and events -- through a sequential pipeline. Each component, together with its parameters, is expressed as a symbolic abstract representation of the time series. This program-friendly representation, along with a natural language description of each component and an instruction $q$, is then passed to an LLM for downstream tasks. For editing, the LLM outputs an edited structural program representation, which is then reconstructed back into a time series, whereas for captioning and reasoning, it produces a natural language response.}
\label{fig:overview}
\vspace{-10pt}
\end{center}
\end{figure*}

\section{Time-series-to-Structured-Program}
\label{sec:decomposition}

\name is a representation method that decomposes a time series into structural and interpretable components -- trend, periodicity, and events -- and expresses them in a symbolic abstract representation, enabling LLMs to reason over decomposed temporal structure rather than raw time-series values. We first formalize the notation and problem setup, then describe how each component is extracted from a raw time series through a deterministic decomposition pipeline. Building on this formulation, we show that the original time series can be faithfully reconstructed from the resulting symbolic representation, demonstrating that the decomposition preserves the essential temporal information. Finally, we present the structured program syntax used as the interface between time series and LLMs.

\define{Notations} Let $T\!\in\!\mathbb{R}^+$ be the time horizon where we assume each time series originates from some true underlying continuous trajectory $x\!:\![0, T]\!\rightarrow\!\mathbb{R}$, which is observed at $N$ discrete time points $\{t_1, \dots, t_N \} \subset [0,T]$. This produces a discrete observation vector $\mathbf{x}\!=\![x(t_1), ...,x(t_N)]^{\top}\!\in\!\mathbb{R}^{N}$. 
An LLM, $f_\theta$, performs a time-series reasoning task by taking a natural language instruction $q \in \mathcal{Q}$ together with a time series $\mathbf{x}$ as input, and producing an output 
\begin{equation}
\label{eq:ts_reasoning}
y = f_\theta(q, \mathbf{x})\in\mathcal{Y}.
\end{equation}
Here, $\mathcal{Y}$ denotes the space of natural-language outputs. Depending on the instruction specified by $q$, the model output $y \in \mathcal{Y}$ may correspond to either: (i) a textual response, such as time-series \textit{captioning} and \textit{question answering} or (ii) a structural representation of a time series that can be decoded back to a reconstructed time series $\hat{\mathbf{x}}\!\in\!\mathbb{R}^{N}$, as in time-series \textit{editing}. For simplicity, we focus on the univariate setting throughout this work.

\define{Problem Formulation} Our central question is how to transform a raw time series $\mathbf{x}$ into a representation that enables $f_\theta$ to effectively reason and, therefore, produce a faithful output $y$ in~Eq.~\eqref{eq:ts_reasoning}. 
The current dominant approach~\cite{xie2025chatts, sen2025bedtime} represents time series as a serialized string, \ie $\texttt{str}(\mathbf{x})$, where the sequence is expressed as a list of numerical values. While this representation is straightforward and preserves the full fidelity of the original time series, it presents two major challenges for LLMs. First, the tokenization cost grows linearly with both the sequence length~$N$ and the numerical precision of the value~\cite{sen2025bedtime}. Second, the model must implicitly infer high-level temporal structure (\ie trend, periodic patterns, and salient events) directly from raw numerical sequences, a task known to be difficult for LLMs~\cite{zhou2025fone}. 
These limitations suggest that the key bottleneck lies not only in the model, but also in the representation itself. Motivated by this observation, we propose a deterministic, training-free, and invertible transformation

\begin{equation} \label{eq:representation}
    \phi: \mathbb{R}^{N} \rightarrow \mathcal{P}, \quad \phi^{-1}(\phi(\mathbf{x})) = \mathbf{x},
\end{equation}
which maps a raw time series $\mathbf{x}$ to a structured program representation $\phi(\mathbf{x})\in\mathcal{P}$, where~$\mathcal{P}$ denotes the space of structured programs. Under this formulation, the time-series reasoning task now becomes: $y = f_\theta(q, \phi(\mathbf{x}))$. In the following sections, we describe how $\phi$ is designed to preserve the essential temporal structure of $\mathbf{x}$ while producing a representation that aligns naturally with the reasoning capabilities of LLMs.

\subsection{Structural Decomposition}
We design $\phi$ as a structural decomposition of $\mathbf{x}$ into three components: trend, periods, and events. This design is motivated by two key observations. First, these components together capture much of the essential structure of a time series: The trend captures the overall trajectory, period explains the recurring temporal patterns, and events characterize abrupt local deviations. 
Together, they provide a compact yet expressive description of a time series. Moreover, the original time series can be reconstructed by combining these components through element-wise summation, ensuring that the decomposition preserves the underlying temporal information.

Second, structural decomposition is a well-established principle in both classical~\cite{cleveland1990stl} and modern time-series literature~\cite{wu2021autoformer, kim2025transpl}, providing a strong foundation for extracting each component using existing methodologies. Formally, we write the decomposition of $\mathbf{x}$ as follows: 
\begin{equation}
\label{eq:decomposition}
    \mathbf{x} = \boldsymbol{\tau}+ \sum_{k=1}^{K_p} \boldsymbol{\pi}_k + \sum_{j=1}^{K_e} \boldsymbol{\epsilon}_j + \boldsymbol{r},
\end{equation}
where $\boldsymbol{\tau},\boldsymbol{\pi}_k,\boldsymbol{\epsilon}_j,\boldsymbol{r}\!\in\!\mathbb{R}^{N}$ are trend, period, event, and the residual, respectively and $K_p$ and $K_e$ are the number of periodic and event components. We now describe how each component is obtained through a deterministic and sequential pipeline. 

\define{Trend} The trend represents the general direction of how the time series is moving. Here, we model $\boldsymbol{\tau}$ as a composition of degree-$d$ B-spline functions
\begin{equation}
\label{eq:trend}
    \tau(t) = \sum_{i=1}^{M} c_i\, B_{i,d}(t; \boldsymbol{\kappa}),
\end{equation}
where $\boldsymbol{\kappa}\!=\![k_1,\dots, k_M]$ is a vector of knots and~$\{c_i\}_{i=1}^{M}$ are the coefficients for the B-spline functions~\cite{de1972calculating}. The discrete trend vector is then obtained by evaluating $\tau(t)$ at the observed time points, \ie $\boldsymbol{\tau}\!=\![\tau(t_1),\dots,\tau(t_N)]^\top$. 

B-splines are smooth and locally adaptive, making them suitable for trend modeling, and their knot-coefficient parameters make them interpretable. The knots and coefficients for $\tau(t)$ are obtained by solving a smoothness-penalized least-squares problem. When domain knowledge is available, the knots can be fixed~(\eg uniform grid), with only the coefficients fitted to the data.

\define{Periods} The periodic components capture recurring patterns in the time series, such as seasonal patterns. Using the detrended signal $\mathbf{x}-\boldsymbol{\tau}$, we model each periodic component $\pi_k$ as a sinusoid:
\begin{equation*}
\label{eq:periodic}
    \pi_k(t) = a_k \sin(2\pi \omega_k t) + b_k \cos(2\pi \omega_k t),
\end{equation*}
parametrized by frequency $\omega_k$ and basis coefficients $a_k$ and $b_k$. We select the dominant peaks~(up to $K_p$ components) from the Fourier spectrum and use them as $\omega_k$, where least squares is used to fit the basis coefficients. While the above formulation assumes global periodic components, time series can also have varying periodic structures between time points. As such, we also allow sinusoids to be fitted piecewise within each segment defined by the trend knot positions. These structural components are explicitly encoded in our \name representation.

\define{Events} Events represent abrupt deviations from the smooth trend and periods. Working on the residual value after subtracting the trend and periodic components, we model two different types of events: spikes and Gaussian. 
\begin{equation*}
\label{eq:event}
    \epsilon_j(t) =
    \begin{cases}
        a_j\, \delta(t - \mu_j), & \text{Spike,} \\[2pt]
        a_j\, \exp\!\left(-\dfrac{(t - \mu_j)^2}{2\sigma_j^2}\right), & \text{Gaussian,}
    \end{cases}
\end{equation*}
where $\mu_j, a_j, \sigma_j$ denote the event time, amplitude, and width, respectively, and $\delta(\cdot)$ is the Dirac-delta function. Spikes are used for instantaneous and extreme changes (\eg a shock), while Gaussian events are for smooth but notable changes that are not reflected in both trend and periods. 

\define{Residuals or Noise} After removing the trend, periods, and events from the original signal, what remains is a small, noise-like residual. Depending on the task, we either retain the full residual values outside of the program for exact reconstruction of the original series (\eg editing tasks that require returning a full time series), or parametrize it as a  Gaussian, which is included as part of the program representation (\eg captioning or question answering, where exact reconstruction is unnecessary).

\subsection{Symbolic Abstract Representation}
Building on the structural decomposition, we now aggregate the components and parameters into our symbolic abstract representation of time series~(shown in~\cref{fig:overview}, full detail in~Appendix~\ref{appendix:full_t2sp}), and pass it to an LLM for downstream tasks. Unlike string representation that lists value-by-value, our \name representation makes each of the structural components \emph{explicit}, paired with a natural-language description for each component, and task instruction $q$. This design reduces the burden on the LLM to infer structure from numbers alone, where the LLM can now reason directly over the components. Moreover, as shown in Eq.~\eqref{eq:decomposition}, we can reconstruct the original time series by adding each component, making our representation invertible. To summarize, given a raw time series, \name deterministically decomposes it into its structural components, converts them into a symbolic abstract representation, and passes it to an LLM along with a task-specific instruction. The resulting representation is training-free, invertible, and compatible with any off-the-shelf LLMs. 
\section{Experiments}
\begin{table*}[!ht]
\centering
\footnotesize
\setlength{\tabcolsep}{6pt}
\renewcommand{\arraystretch}{1.1}
\resizebox{0.9\textwidth}{!}{%
\begin{tabular}{l|ccc|ccc|ccc|ccc}
\toprule
\multirow{2}{*}{Method}
& \multicolumn{3}{c|}{\textbf{Trend}}
& \multicolumn{3}{c|}{\textbf{Periodic}}
& \multicolumn{3}{c|}{\textbf{Event}}
& \multicolumn{3}{c}{\textbf{Average}} \\
& Fid. $\red\uparrow$ & Pre.$\red\uparrow$ & Succ.$\red\uparrow$
& Fid.$\red\uparrow$ & Pre.$\red\uparrow$ & Succ.$\red\uparrow$
& Fid.$\red\uparrow$ & Pre.$\red\uparrow$ & Succ.$\red\uparrow$
& Fid.$\red\uparrow$ & Pre.$\red\uparrow$ & Succ.$\red\uparrow$ \\
\midrule
\multicolumn{13}{l}{\footnotesize\texttt{GPT-5.4}} \\
\hspace{1.5em} Raw
& 0.829 & 0.589 & 0.925
& 0.653 & 0.562 & 0.938
& 0.776 & 0.911 & 0.975
& 0.753 & 0.687 & 0.946 \\
\hspace{1.5em} Vision
& 0.452 & 0.389 & 0.875
& 0.489 & 0.391 & 0.800
& 0.538 & 0.270 & 0.875
& 0.493 & 0.350 & 0.850 \\
\rowcolor{imprcolor}
\hspace{1.5em} \name (Ours)
& \textbf{0.976} & \textbf{0.878} & \textbf{1.000}
& \textbf{0.866} & \textbf{0.791} & \textbf{1.000}
& \textbf{0.880} & \textbf{0.984} & \textbf{1.000}
& \textbf{0.907} & \textbf{0.884} & \textbf{1.000} \\

\cmidrule(lr){1-13}
\multicolumn{13}{l}{\footnotesize\texttt{Claude-haiku-4.5}} \\
\hspace{1.5em} Raw
& 0.377 & 0.668 & 0.963
& 0.495 & 0.653 & \textbf{1.000}
& 0.718 & 0.904 & 0.988
& 0.530 & 0.742 & 0.983 \\
\hspace{1.5em} Vision
& 0.438 & 0.198 & \textbf{0.988}
& 0.536 & 0.175 & \textbf{1.000}
& 0.462 & 0.000 & 0.988
& 0.479 & 0.124 & 0.992 \\
\rowcolor{imprcolor}
\hspace{1.5em} \name (Ours)
& \textbf{0.844} & \textbf{0.862} & \textbf{0.988}
& \textbf{0.866} & \textbf{0.791} & \textbf{1.000}
& \textbf{0.880} & \textbf{0.984} & \textbf{1.000}
& \textbf{0.863} & \textbf{0.879} & \textbf{0.996} \\

\cmidrule(lr){1-13}
\multicolumn{13}{l}{\footnotesize\texttt{Gemini-3.1-flash-lite}} \\
\hspace{1.5em} Raw
& 0.509 & 0.582 & 0.912
& 0.509 & 0.615 & 0.988
& 0.732 & 0.926 & \textbf{1.000}
& 0.583 & 0.708 & 0.967 \\
\hspace{1.5em} Vision
& 0.312 & 0.391 & 0.762
& 0.482 & 0.259 & 0.750
& 0.470 & 0.272 & 0.738
& 0.421 & 0.307 & 0.750 \\
\rowcolor{imprcolor}
\hspace{1.5em} \name (Ours)
& \textbf{0.971} & \textbf{0.877} & \textbf{1.000}
& \textbf{0.868} & \textbf{0.791} & \textbf{1.000}
& \textbf{0.867} & \textbf{0.967} & 0.975
& \textbf{0.902} & \textbf{0.879} & \textbf{0.992} \\

\cmidrule(lr){1-13}
\multicolumn{13}{l}{\footnotesize\texttt{Qwen-3.5-9B}} \\
\hspace{1.5em} Raw
& 0.564 & 0.424 & 0.725
& 0.469 & 0.513 & 0.838
& 0.595 & 0.760 & 0.812
& 0.543 & 0.566 & 0.792 \\
\hspace{1.5em} Vision
& 0.478 & 0.204 & 0.762
& 0.727 & 0.406 & 0.950
& 0.639 & 0.073 & \textbf{1.000}
& 0.615 & 0.228 & 0.904 \\
\rowcolor{imprcolor}
\hspace{1.5em} \name (Ours)
& \textbf{0.935} & \textbf{0.847} & \textbf{0.963}
& \textbf{0.866} & \textbf{0.791} & \textbf{1.000}
& \textbf{0.874} & \textbf{0.993} & \textbf{1.000}
& \textbf{0.892} & \textbf{0.877} & \textbf{0.988} \\
\midrule
\multicolumn{13}{l}{\footnotesize\textit{Others}} \\
\hspace{1.5em} ChatTime-7B
& 0.358 & 0.111 & 0.800
& 0.307 & 0.091 & 0.800
& 0.402 & 0.000 & 0.800
& 0.356 & 0.067 & 0.800 \\
\hspace{1.5em} Verbal-TS$^{*}$
& 0.351 & 0.419 & 1.000
& 0.501 & 0.118 & 1.000
& 0.495 & 0.000 & 1.000
& 0.452 & 0.179 & 1.000 \\
\hspace{1.5em} InstructTime$^{*}$
& 0.598 & 0.616 & 1.000
& 0.638 & 0.311 & 1.000
& 0.503 & 0.044 & 1.000
& 0.580 & 0.324 & 1.000 \\
\bottomrule
\end{tabular}
}
\caption{\define{Performance on the TSEdit benchmark} Fid., Pre., and Succ. denote edit fidelity, preservation, and success rate, respectively. We compared our \name with raw time series (Raw) and image~(Vision) representation using the same LLM as a backbone. We also compared with training-based models~(denoted with *), where 6000 samples were used for model training. For failed samples, where the model produced an invalid output format, Fid. and Pre. were assigned a score of zero.} 
\label{tab:editing_results}
\vspace{-5pt}
\end{table*}

We verify the applicability of \name on time-series editing, captioning, and reasoning tasks. As \name works on any off-the-shelf LLMs, we evaluate on three closed-source models accessed via API, ranging from flagship~(\texttt{GPT-5.4})~to lightweight~(\texttt{Claude-haiku-4.5},~\texttt{Gemini-3.1 -flash-lite}), and an open-source model (\texttt{Qwen-3.5-9B}). Within each LLM, we compare three different representations of time series: string (\textbf{raw}), visual rendering (\textbf{vision}), and \name (\textbf{ours}). In addition, we compare against task-specialized baselines, introduced in each task's subsection. We provide the details of all datasets in~Appendix~\ref{appendix:datasets}.

\subsection{Time-series Editing}
A key advantage of representing time series as structural components is that it enables \emph{targeted editing}. When time series are represented as raw sequences or images, the LLM must implicitly reason over individual values or pixels, making localized modifications difficult. In contrast, \name exposes the time series as a symbolic abstract representation that LLMs can manipulate directly~\cite{gao2023pal}. This allows the model to modify only the intended component while preserving the remaining structure, providing precise edits in a token-efficient manner. Moreover, since edits are performed directly on symbolic components, the modifications made by the LLM (\eg modification, insertion, deletion) remain human-interpretable.



\define{Setup} We evaluate the editing capability of \name on two different tasks. First, we construct and evaluate on \textbf{TSEdit}, a synthetic benchmark consisting of time series paired with natural-language editing instructions targeting trend, periodic, or event components (\eg \textit{Remove the anomaly at $t=50$}), along with a corresponding ground-truth edited time series. Since the synthetic data are generated by independently sampling and composing trend, periodic, and event components, the ground-truth edits come directly from the data generation process and are independent of our decomposition method. Second, we evaluate on ETTh1~\cite{wu2021autoformer}, a real-world oil temperature forecasting dataset, with editing instructions that mimic the unstructured, ambiguous phrasing of real users. For instance, instruction such as \textit{``Remove the sudden drop around the center''} leaves the notion of \textit{center} open to interpretation. For this task, we use human evaluation because component-level ground truth is not well defined for real-world time series. The key metrics for both tasks are: \textit{fidelity}~(\ie performing the targeted edit) and \textit{preservation}~(\ie leaving the other components intact). In addition, we report the \textit{success rate}, measuring how often the LLM produces a response that conforms to the required output format. In addition to the LLM-based methods, we compare with a time-series language model, ChatTime-7B~\cite{wang2025chattime}; a text-conditioned time-series generative model, Verbal-TS~\cite{gu2025verbalts}; and the state-of-the-art editing model, InstructTime~\cite{qiu2026instruction}.
\begin{figure}[t]
\begin{center}
\center{\includegraphics[width=0.9\columnwidth]{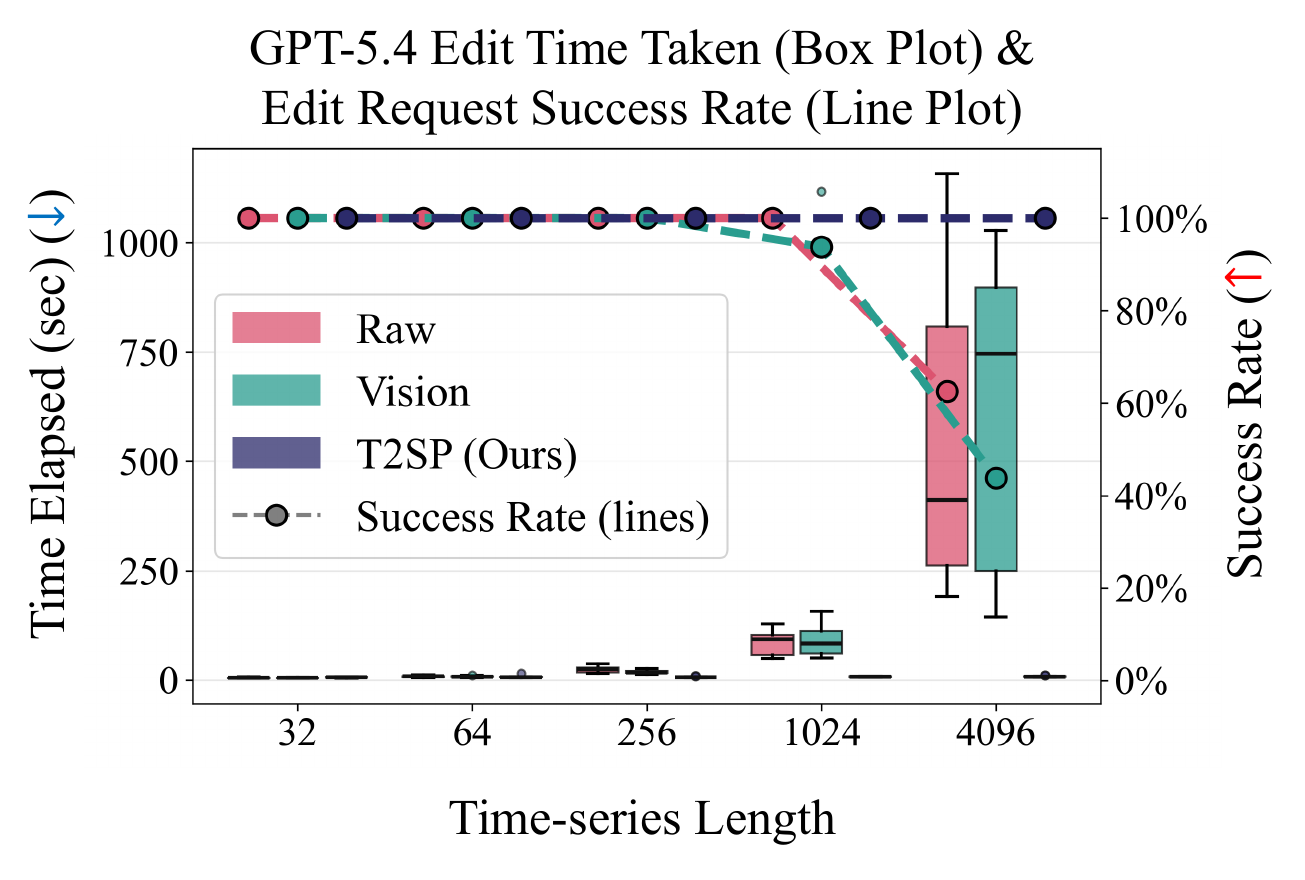}}
\vspace{-10pt}
    \caption{\define{Time Taken and Success Rate Across Sequence Length} The left and right axes represent the time taken (seconds) to perform the editing and the success rate, respectively. The three representations perform comparably up to a sequence length of 256, beyond which the raw and vision-based baselines deteriorate rapidly, while \name remains stable.}
    \label{fig:time_taken}
\end{center}
\vspace{-10pt}
\end{figure}


\define{TSEdit Results} \cref{tab:editing_results} shows that \name consistently achieves the strongest performance across all component types (\ie trend, periodic, and event edits) and across all LLM backbones. Overall, editing performance improves with model capability for every representation, with \texttt{GPT-5.4} performing best, followed by smaller models such as \texttt{Gemini}, \texttt{Claude}, and \texttt{Qwen}. This trend highlights the importance of being able to leverage highly capable closed-source LLMs without additional task-specific training. A notable observation is that \name achieves a substantially better balance between fidelity and preservation than competing representations, which often improve one metric at the expense of the other. We attribute this behavior to the explicit structural decomposition provided by our symbolic abstraction: because each component is independently represented, the LLM can precisely modify the targeted component while leaving the remaining structure untouched. Importantly, \name achieves these gains in a fully training-free manner, whereas models such as Verbal-TS and InstructTime are extensively trained on synthetic editing samples constructed from combinations of the same underlying editing operations.

\begin{table}[t]
\centering
\footnotesize
\setlength{\tabcolsep}{6pt}
\renewcommand{\arraystretch}{0.9}
\begin{tabular}{lccc}
\toprule
Human Evaluation & Fid. $\red\uparrow$  & Pre. $\red\uparrow$  & Overall $\red\uparrow$ \\
\midrule
\rowcolor{gray!10}\multicolumn{4}{l}{\footnotesize\texttt{GPT-5.4}} \\
\hspace{1.5em} \name\ vs.\ Raw    & 58.9\% & 56.7\% & 60.0\% \\
\hspace{1.5em} \name\ vs.\ Vision & 84.3\% & 91.4\% & 92.9\% \\
\midrule
\rowcolor{gray!10}\multicolumn{4}{l}{\footnotesize\texttt{Claude-haiku-4.5}} \\
\hspace{1.5em} \name\ vs.\ Raw    & 84.4\% & 63.3\% & 80.0\% \\
\hspace{1.5em} \name\ vs.\ Vision & 91.4\% & 97.1\% & 92.9\% \\
\midrule
\rowcolor{gray!10}\multicolumn{4}{l}{\footnotesize\texttt{Qwen3.5-9b}} \\
\hspace{1.5em} \name\ vs.\ Raw    & 79.2\% & 73.6\% & 79.2\% \\
\hspace{1.5em} \name\ vs.\ Vision & 88.6\% & 100.0\% & 88.6\% \\
\bottomrule
\end{tabular}
\caption{\define{Pairwise Evaluation on ETTh1} Each row compares \name against a baseline; values denote the Win Rate~(\%) of \name over the baseline. Fid. and Pre. denotes the fidelity and preservation. Overall is the final preference. Detailed Win/Tie/Lose rates are provided in Appendix~\ref{appendix:humaneval}.}
\label{tab:pairwise_comparison}
\end{table}
\define{Human Evaluation} We further conduct a pairwise human evaluation on the real-world ETTh1 dataset. We recruit ten annotators, each of whom is presented with an original time series, an editing instruction, and two edited outputs presented in randomized order without method labels. One time series is produced by \name and the other by either the raw or vision-based baseline. Annotators are then asked to select the edited time series that better satisfies the editing instruction. We report the win rate of \name in~\cref{tab:pairwise_comparison}; full details and qualitative examples are provided in Appendix~\ref{appendix:humaneval}. Win rates above 50\% indicate that annotators consistently preferred outputs generated by \name over the competing representations. The performance gap is particularly large against the vision-based baseline, suggesting that image-based representations are poorly suited for editing tasks that require localized and structurally coherent modifications.

\define{Time Taken and Success Rate} We analyze the time taken to perform the editing task and the success rate based on the sequence length in~\cref{fig:time_taken}. We observe that raw and vision-based approaches scale poorly in both runtime and success rate as sequence length increases. We conjecture that these representations incur input and output costs that scale linearly with sequence length, increasing both the reasoning cost and the likelihood of malformed outputs. In contrast, \name remains robust, as even long time series can be faithfully expressed with our symbolic abstract representation.

\begin{table}[t]
\centering
\footnotesize
\setlength{\tabcolsep}{3.7pt}
\renewcommand{\arraystretch}{0.6}
\begin{tabular}{c|c|cc|cc}
\toprule
\multirow{2}{*}{N-shot} & \multirow{2}{*}{Method}
& \multicolumn{2}{c|}{\textbf{Wafer}}
& \multicolumn{2}{c}{\textbf{ECG200}} \\
& & Acc. $\red\uparrow$ & F1 $\red\uparrow$
& Acc. $\red\uparrow$ & F1 $\red\uparrow$ \\
\midrule
\multirow{3}{*}{\textit{zero-shot}}
& Raw
& 0.465 & 0.360
& 0.560 & 0.476 \\
& Vision
& 0.495 & 0.363
& 0.570 & 0.402 \\
& \cellcolor{imprcolor}\name (Ours)
& \cellcolor{imprcolor}\textbf{0.505} & \cellcolor{imprcolor}\textbf{0.368}
& \cellcolor{imprcolor}\textbf{0.660} & \cellcolor{imprcolor}\textbf{0.587} \\
\cmidrule(lr){1-6}
\multirow{3}{*}{\textit{1-shot}}
& Raw
& 0.563 & 0.544
& 0.640 & 0.506 \\
& Vision
& 0.415 & 0.405
& 0.630 & 0.432 \\
& \cellcolor{imprcolor}\name (Ours)
& \cellcolor{imprcolor}\textbf{0.620} & \cellcolor{imprcolor}\textbf{0.611}
& \cellcolor{imprcolor}\textbf{0.670} & \cellcolor{imprcolor}\textbf{0.526} \\
\cmidrule(lr){1-6}
\multirow{3}{*}{\textit{3-shot}}
& Raw
& 0.630 & 0.617
& 0.620 & 0.382 \\
& Vision
& 0.495 & 0.412
& 0.670 & 0.474 \\
& \cellcolor{imprcolor}\name (Ours)
& \cellcolor{imprcolor}\textbf{0.835} & \cellcolor{imprcolor}\textbf{0.835}
& \cellcolor{imprcolor}\textbf{0.690} & \cellcolor{imprcolor}\textbf{0.634} \\
\cmidrule(lr){1-6}
\multirow{3}{*}{\textit{5-shot}}
& Raw
& 0.690 & 0.689
& 0.630 & 0.469 \\
& Vision
& 0.520 & 0.417
& 0.590 & 0.516 \\
& \cellcolor{imprcolor}\name (Ours)
& \cellcolor{imprcolor}\textbf{0.850} & \cellcolor{imprcolor}\textbf{0.849}
& \cellcolor{imprcolor}\textbf{0.710} & \cellcolor{imprcolor}\textbf{0.596} \\
\cmidrule(lr){1-6}
\multirow{3}{*}{\textit{20-shot}}
& Raw
& 0.650 & 0.633
& 0.650 & 0.442 \\
& Vision
& 0.560 & 0.546
& 0.650 & 0.442 \\
& \cellcolor{imprcolor}\name (Ours)
& \cellcolor{imprcolor}\textbf{0.895} & \cellcolor{imprcolor}\textbf{0.894}
& \cellcolor{imprcolor}\textbf{0.730} & \cellcolor{imprcolor}\textbf{0.668} \\
\bottomrule
\end{tabular}
\caption{\define{Caption-based Classification}
For each representation, captions are first generated from the time series, and classification is performed using only the generated captions. Acc. and F1 denote accuracy and macro-F1, respectively.}
\label{tab:caption_downstream_gemini}
\end{table}
\subsection{Time-series Captioning }

Time-series captioning is the task of producing a natural language description of a time series, where this caption can be used for multi-modal learning, and downstream reasoning~\cite{langer2025opentslm}. A meaningful caption, in this role, must preserve the structural content of the time series with sufficient fidelity, such that the underlying patterns remain recoverable from the description alone.

\define{Setup} To assess the utility of \name for captioning, we adopt a \emph{reference-free} evaluation based on downstream classification performance. Such evaluation can avoid the stylistic biases inherent in human-written reference captions~\cite{hessel2021clipscore}. Specifically, we generate a natural-language caption for each time series and use a separate LLM to classify the sample using only its caption. For this task, we utilize the Wafer and ECG200 datasets~\cite{dau2019ucr}, where classes are distinguished by fine-grained temporal structures rather than by the holistic shape of the time series. We also evaluate a few-shot setup, where we provide $1, 3, 5,$ and $20$ samples for each class as a reference.

\define{Caption Results} Table~\ref{tab:caption_downstream_gemini} reports the classification results using the \texttt{Gemini-3.1-flash-lite} as both the caption generator and classifier for all representations. We report additional model results and qualitative examples in Appendix~\ref{appendix:ts_caption_full}. \name achieves the best accuracy and F1-score across both datasets and few-shot setups. Notably, on Wafer, the gain from additional samples is particularly pronounced for \name -- F1 rises from 0.368 to 0.835 with just three shots, whereas the corresponding gains for vision representation remain modest (vision: 0.363 to 0.412). We attribute this to the nature of a vision-based approach: while the model is suitable in capturing the holistic shape of a time series, they often miss fine-grained temporal structures such as periodicity or localized events. In contrast, \name exposes trend, periodicity, and events as explicit input components, allowing the generated captions to cover both the holistic shape and the detailed temporal structure of the time series.

\subsection{Time-series Reasoning}
\begin{table}[t]
\centering
\small
\setlength{\tabcolsep}{4pt}
\renewcommand{\arraystretch}{0.9}
\resizebox{\columnwidth}{!}{%
\begin{tabular}{l|cc|cc|cc}
\toprule
\multirow{2}{*}{Method}
& \multicolumn{2}{c|}{\textbf{TSQA}}
& \multicolumn{2}{c|}{\textbf{TRQA}}
& \multicolumn{2}{c}{\textbf{ETI}} \\
& ACC $\red\uparrow$ & F1 $\red\uparrow$
& ACC $\red\uparrow$ & F1 $\red\uparrow$
& ACC $\red\uparrow$ & F1 $\red\uparrow$ \\
\midrule
Random Guess
& 0.333 & 0.333
& 0.369 & 0.322
& 0.250 & 0.250 \\
\midrule
\multicolumn{7}{l}{\footnotesize\texttt{GPT-5.4}} \\
\hspace{1.5em} Raw
& 0.818 & 0.645
& 0.740 & 0.851
& 0.680 & 0.809 \\
\hspace{1.5em} Vision
& 0.810 & 0.646
& 0.710 & 0.830
& \textbf{0.880} & \textbf{0.936} \\
\rowcolor{imprcolor}\hspace{1.5em} \name (Ours)& \textbf{0.857} & \textbf{0.686}
& \textbf{0.815} & \textbf{0.898}
& 0.840 & 0.909 \\
\midrule
\multicolumn{7}{l}{\footnotesize\texttt{Claude-haiku-4.5}} \\
\hspace{1.5em} Raw
& 0.758 & 0.599
& 0.700 & 0.824
& 0.565 & 0.720 \\
\hspace{1.5em} Vision
& 0.732 & 0.580
& 0.750 & 0.857 
& 0.685 & 0.813 \\
\rowcolor{imprcolor}\hspace{1.5em} \name (Ours)& \textbf{0.852} & \textbf{0.681}
& \textbf{0.760} & \textbf{0.864}
& \textbf{0.910} & \textbf{0.955} \\
\midrule
\multicolumn{7}{l}{\footnotesize\texttt{Gemini-3.1-flash-lite}} \\
\hspace{1.5em} Raw
& 0.768 & 0.610
& 0.730 & 0.844
& 0.705 & 0.818 \\
\hspace{1.5em} Vision
& 0.766 & 0.608
& 0.725 & 0.719
& 0.925 & 0.961 \\
\rowcolor{imprcolor}\hspace{1.5em} \name (Ours)& \textbf{0.859} & \textbf{0.683}
& \textbf{0.760} & \textbf{0.864}
& \textbf{0.930} & \textbf{0.962} \\
\midrule
\multicolumn{7}{l}{\footnotesize\texttt{Qwen-3.5-9B}} \\
\hspace{1.5em} Raw
& 0.745 & 0.579
& 0.735 & 0.847
& 0.815 & 0.901 \\
\hspace{1.5em} Vision
& 0.743 & 0.591
& \textbf{0.755} & \textbf{0.860}
& 0.885 & 0.939 \\
\rowcolor{imprcolor}\hspace{1.5em} \name (Ours)& \textbf{0.867} & \textbf{0.694}
& 0.727 & 0.842
& \textbf{0.950} & \textbf{0.974} \\
\midrule
ChatTime-7B
& 0.521 & 0.516
& 0.314 & 0.245
& 0.220 & 0.100 \\
\bottomrule
\end{tabular}%
}
\caption{Performance comparison across TSQA, TRQA, and ETI tasks. Best scores in \textbf{bold}.}
\label{tab:tsqa_results}
\end{table}

Time-series reasoning is the task of answering a question about a given time series, typically formulated as a multiple-choice question where the LLM selects the correct answer from a set of candidates~\cite{kong2025time, wang2025chattime}. Compared to captioning, reasoning provides a stricter test of structural understanding, as each answer is judged against a ground truth. This makes reasoning a suitable testbed for \name, where we can evaluate whether the LLM can effectively reason over our symbolic abstract representations.

\define{Setup} We evaluate on three time-series reasoning benchmarks that span complementary question types: TSQA~\cite{wang2025chattime} asks whether a time series satisfies a specified structural pattern, while ETI~\cite{merrill2024language} and TRQA~\cite{jingtrqa} require selecting the explanation that best matches a given time series. We compare \name against raw and vision-based representations and to ChatTime-7B~\cite{wang2025chattime}.

\define{Results} In~\cref{tab:tsqa_results}, \name is the best performing representation in 10 out of 12 cases. The performance gain is especially notable in both the TSQA and ETI tasks, which require reasoning over the temporal structures of the time series (\eg \textit{which structure has a linearly increasing trend}). In particular, for the \texttt{Claude} model on the ETI task, the raw representation achieves only 56.5\% accuracy, whereas \name reaches 91.0\%, demonstrating a substantial performance gap of over 34 percentage points. We also note that vision-based representation is a competitive baseline, an observation that aligns with existing works~\cite{sen2025bedtime}. In fact, the two cases where \name underperforms are both lost to vision. This observation suggests that existing benchmarks may, to some extent, be solvable through purely visual cues, pointing to the need for the development of benchmarks that require more explicit reasoning over the structural properties of time series rather than those that can be solved by visual inspection.
\section{Conclusion}
In this work, we propose \name, a novel time-series representation method for LLM interfacing. Specifically, \name decomposes a time series into structural components -- trend, periods, and salient events -- and aggregates them into a structured symbolic representation that is re-expressed in a program-friendly form. Then, the LLM reasons over this structured representation space instead of the raw numerical series. As such, \name aligns the input with the symbolic and code-like modalities LLMs were natively pretrained on, allowing them to apply their existing reasoning capabilities directly. Notably, our \name representation is deterministic, training-free, and compatible with any off-the-shelf LLM, including the strongest performing closed-source models accessible only via APIs. We demonstrate \name's capability on time-series editing, caption generation, and reasoning, establishing it as an effective interface that connects time series and the reasoning capabilities of LLMs.

\section*{Limitations}
In this work, we proposed \name to represent a time series as a symbolic abstract representation -- composed of trend, periods, and events -- that works as an effective interface between time series and LLMs. While our decomposition faithfully captures the structural composition of a time series, the trend--period--event decomposition may not be the most suitable abstraction for every time-series task. For instance, electrocardiogram (ECG) signals are characterized by domain-specific morphological primitives such as the P, Q, R, S, and T waves, whose clinical meaning is tied to their precise shape and relative timing rather than to a global trend or periodicity. Such fine-grained, domain-specific structures are not directly captured by our trend--period--event decomposition, and a representation tailored to these primitives would likely be more effective. Therefore, we acknowledge that our decomposition strategy may not be universally optimal across all tasks and datasets. Our goal in this paper is not to argue that our decomposition is the optimal strategy, but rather to demonstrate the potential of structural, program-friendly representations as an effective interface between time series and LLMs. We view the design of domain-specific decompositions as a promising direction for future work.

\section*{Ethical considerations}
This work introduces a representation method for time-series analysis with large language models. The contribution is methodological. Also, our experiments are conducted on both a synthetic dataset (explained in~\cref{appendix:tseditdata}) and publicly available time-series benchmarks. We do not collect new data from human subjects beyond the human evaluation described in~\cref{appendix:humaneval}, for which annotators were recruited with informed consent and the evaluation involved no sensitive or personally identifying information. We do not foresee direct ethical concerns or potential for misuse beyond those generally associated with applying large language models to numerical and temporal data.

\label{sec:bibtex}


\bibliography{custom}
\appendix
\newpage
\hfill
\newpage
\section{Full \name Representation}\label{appendix:full_t2sp}
We provide a visual illustration of the full \name representation for a given time series. Following the deterministic decomposition pipeline described in~\cref{sec:decomposition}, a raw series is decomposed into its structural components -- trend, periods, and events -- with the residual either retained for exact reconstruction or parametrized as a Gaussian, depending on the downstream task.

\begin{tsprogrambox}[title={An example of \name representation}]
Series(
    trend=BSplineTrend(
        degree=1, smoothness=5, knot_mode='fixed',
        uniform_knots=[0, 23.5, 47],
        knot_ts_values=[0.2326, 3.853, 8.5785]
    ),
    periodic=[
        Sinusoid(amplitude=1.3278, period=12, phase=0.5414, r2=0.226),
        Sinusoid(amplitude=0.6435, period=4, phase=-1.9744, r2=0.0686)
    ],
    events=[
        SpikeEvent(time=10, amplitude=6.1962),
        GaussianEvent(time=35, width=1.4943, amplitude=3.6144)
    ],
    noise=Residual(std=1.0511, mean=-0.4111)
)
\end{tsprogrambox}
\vspace{-0.5em}
\captionof{figure}{
\textbf{Example of \name representation.}
We format the trend, events, and periods into a structured symbolic abstract representation. This representation is human-interpretable and invertible.
}
\label{fig:ts2p-example}

\begin{figure}[!ht]
\begin{center}\center{\includegraphics[width=\columnwidth]{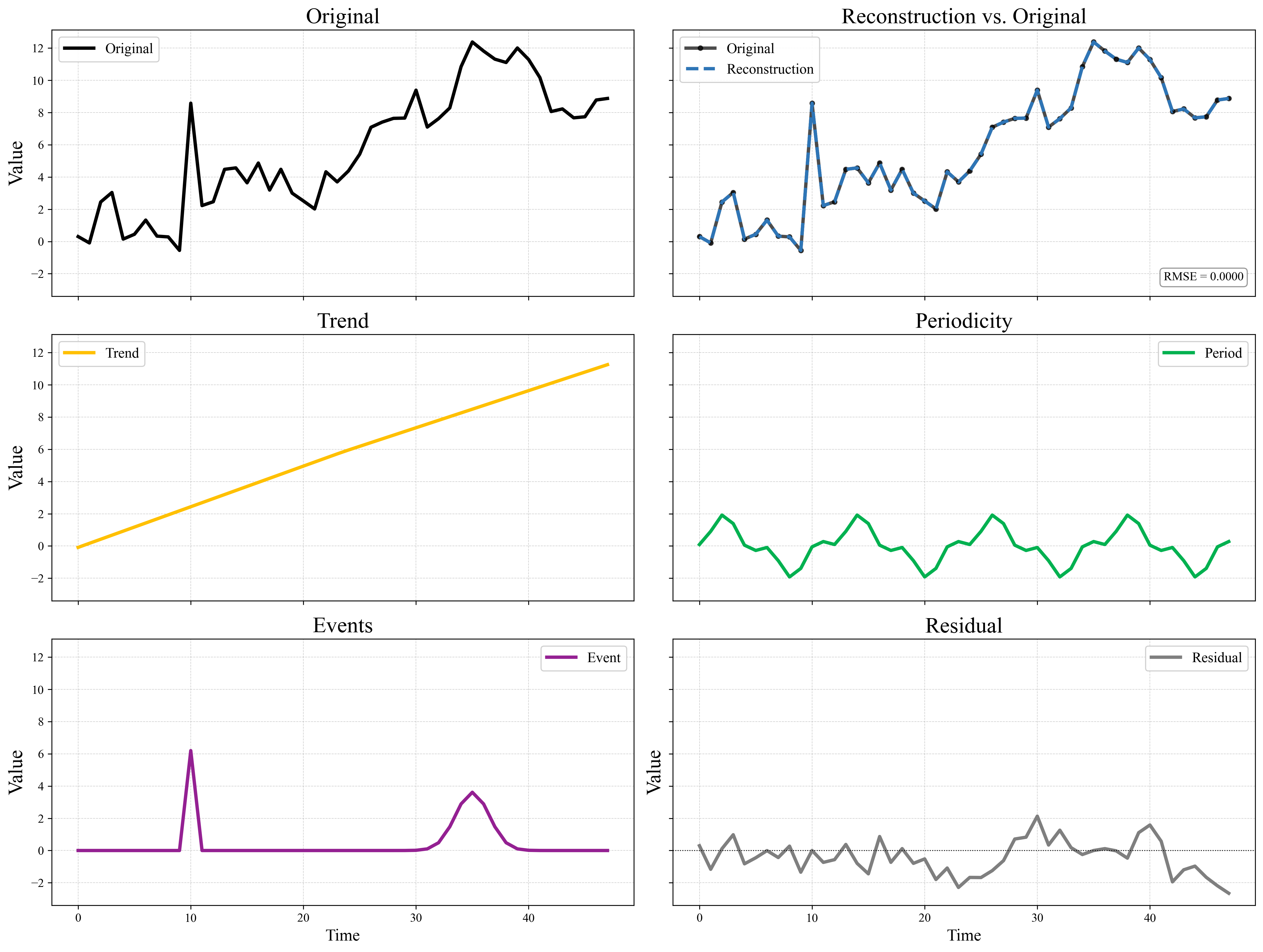}}
\caption{\define{Decomposition from \name} We provide a visual illustration of the decomposition.}
\label{appendix_fig:t2sp}
\end{center}
\end{figure}

\hfill
\newpage
\section{Datasets \& Baseline Implementation}
\label{appendix:datasets}
\begin{table*}[!ht]
\centering
\caption{Overview of datasets used in this paper}
\label{tab:datasets}
\resizebox{\textwidth}{!}{%
\begin{tabular}{llcccl}
\toprule
Dataset & Task & \# Train & \# Test & Sequence Length & Note \\
\midrule
\rowcolor{gray!10}\multicolumn{6}{l}{\textit{Editing}} \\
\midrule
TSEdit-Trend  & Editing & \multirow{3}{*}{6{,}000} & 80 & $\{32, 64, 256, 1024, 4096\}$ & 16 test samples per length \\
TSEdit-Period & Editing &                          & 80 & $\{32, 64, 256, 1024, 4096\}$ & We used training samples only to train the baseline models  \\
TSEdit-Event  & Editing &                          & 80 & $\{32, 64, 256, 1024, 4096\}$ & (InstructTime, Verbal-TS), which require training for time-series editing.  \\
\midrule
ETTh1 & Editing & --      & 15 & 1440 & Human evaluation; 5 curated samples each for Trend, Period, Event \\
\midrule
\rowcolor{gray!10}\multicolumn{6}{l}{\textit{Captioning}} \\
\midrule
Wafer  & Classification & 6{,}164 & 200 & --  & \multirow{2}{=}{Train samples used as few-shot examples. } \\
ECG200 & Classification & 100     & 100 & 96  & \\
\midrule
\rowcolor{gray!10}\multicolumn{6}{l}{\textit{Question Answering}} \\
\midrule
TSQA-Trend       & MCQ      & -- & 100 & $\{64, 128, 256, 512\}$ & Stratified sampling by length (100 samples) \\
TSQA-Seasonality & MCQ      & -- & 100 & $\{64, 128, 256, 512\}$ & Stratified sampling by length (100 samples) \\
TSQA-Volatility  & MCQ      & -- & 100 & $\{64, 128, 256, 512\}$ & Stratified sampling by length (100 samples) \\
TSQA-Outliers    & MCQ      & -- & 100 & $\{64, 128, 256, 512\}$ & Stratified sampling by length (100 samples) \\
TRQA             & MCQ, T/F & -- & 200 & $131 \pm 64$            & Randomly sampled (200 samples) \\
ETI              & MCQ      & -- & 200 & $422 \pm 376$           & Randomly sampled (200 samples) \\
\bottomrule
\end{tabular}%
}
\end{table*}

\subsection{TSEdit Dataset}
\label{appendix:tseditdata}
We construct the \textbf{TSEdit} dataset to evaluate instruction-based time-series editing. Prior work on time-series editing~\cite{qiu2026instruction} adopts an \emph{attribute-based} formulation, where edits are framed as adding or removing the presence of a predefined attribute in the series (\eg the existence of bradycardia, a type of abnormal heart event). While this formulation is useful for controlled evaluation, it differs from how humans typically describe edits in practice, where instructions often specify \emph{which} component to change and \emph{how} -- for example, ``remove the anomaly at $t=50$'' or ``flatten the trend''. To better reflect this setting, we construct TSEdit by synthetically generating pairs of free-form natural-language instructions and their corresponding ground-truth target series.

To generate each time series, we first initialize generating functions for the trend, period, and event components, and aggregate them to form the final series. This construction enables us to access the underlying attributes of each component (\eg trend slope, periodicity, event time), and consequently to the ground-truth attributes of the edited series. Importantly, \textbf{we emphasize that each series is \emph{constructed from} its components, rather than the components being inferred from a pre-existing series.} As a result, even our own decomposition does not recover the original generating parameters exactly: the generating functions and our decomposition primitives do not share the same parametrization, so applying our representation to TSEdit is not a trivial identity mapping. This ensures that TSEdit does not artificially favor our representation, and that our evaluation reflects the genuine difficulty of recovering and editing the underlying structure. Below, we explain the type of instruction used for each task. We will publicly release the dataset upon acceptance.

\subsection{Baseline Implementation for TSEdit}
Instruction-based time-series editing has only been recently studied, and to the best of our knowledge, InstructTime~\cite{qiu2026instruction} is the only existing method that directly targets this task. We therefore use InstructTime as the primary editing baseline and follow its original training and inference procedure. In addition, to cover a broader editing paradigm, we consider a time-series generation-based setting from TEdit~\cite{jing2024towards}, where time-series editing is performed by generating a new time series that reflects the desired changes rather than directly modifying the input series. In this setting, the editing instruction specifies the properties of the target output series, and the model generates a series that satisfies those properties from scratch. Based on this perspective, we include Verbal-TS~\cite{gu2025verbalts} as an editing-from-scratch baseline. Since both InstructTime and Verbal-TS require training samples for editing, we independently generate 6000 training samples from the same distribution of TSEdit. We also generate caption pairs that reflect the attributes needed to train InstructTime and Verbal-TS. At inference time, both models generate the edited time series conditioned on the caption of the target series that reflects the requested edit.

\section{Time Series Editing: Human Evaluation}
\label{appendix:humaneval}

\begin{table}[!ht]
\centering
\scriptsize
\caption{Edit operations defined in the TSEdit dataset. Each operation modifies a specific structural component (trend, periodicity, or event) of the input series. $N$ denotes a randomly sampled magnitude parameter.}
\label{tab:edit_operations}
\begin{tabular}{ll}
\toprule
Operation & Description \\
\midrule
\rowcolor{gray!10}\textit{Trend} & \\
\midrule
\texttt{edit\_flatten}         & Flatten the trend to be constant  \\
\texttt{edit\_increase\_slope} & Increase the slope by $+N\%$ \\
\texttt{edit\_decrease\_slope} & Decrease the slope by $-N\%$ \\
\texttt{edit\_reverse}         & Reverse the trend direction \\
\midrule
\rowcolor{gray!10}\textit{Periodic} & \\
\midrule
\texttt{periodic\_remove\_largest}  & Zero out the seasonality \\
\texttt{periodic\_amp\_increase}    & Increase amplitude by $+N\%$ \\
\texttt{periodic\_amp\_decrease}    & Decrease amplitude by $N\%$ \\
\texttt{periodic\_change\_period}   & Change the period by $\pm N\%$ \\
\midrule
\rowcolor{gray!10}\textit{Event} & \\
\midrule
\texttt{event\_remove\_largest} & Remove the largest event \\
\texttt{event\_remove\_second}  & Remove the second largest event \\
\texttt{event\_shift}           & Shift the largest event by $N$ time steps \\
\texttt{event\_reduce\_amp}     & Reduce amplitude largest event to $N\%$ \\
\bottomrule
\end{tabular}
\end{table}

We conduct a pairwise human evaluation to assess whether the edited time series by \name better reflects a given editing instruction compared to alternative representations (\ie raw, vision).

\define{Data curation} For this evaluation, we curated 15 samples of length 1440 (equivalent to a 2-month period) from the real-world ETTh1 dataset. We could not select data points at random, since each human instruction must be matched to a segment that actually exhibits the referenced structure -- for instance, an instruction such as \emph{``remove the anomaly at the center''} requires the corresponding segment to contain an anomalous event at its center. Moreover, since the evaluation requires pairwise comparison across every combination of representation and LLM backbone, this sample size kept the annotation workload manageable within a single one-hour session per annotator, while still being large enough to provide a balanced coverage of the three structural components (5 samples each for trend, period, and event).

\begin{table*}[t]
\centering
\footnotesize
\setlength{\tabcolsep}{4pt}
\renewcommand{\arraystretch}{1.1}
\begin{tabular}{lccc}
\toprule
Baseline
& Fid. (\textbf{Win}/Tie/Lose)
& Pre. (\textbf{Win}/Tie/Lose)
& Overall (\textbf{Win}/Tie/Lose) \\
\midrule

\rowcolor{gray!10}\multicolumn{4}{l}{\textit{GPT-5.4}} \\
\midrule
\hspace{1.5em} \name\ vs.\ Raw    
& \textbf{58.9}\,/\,4.4\,/\,36.7
& \textbf{56.7}\,/\,11.1\,/\,32.2
& \textbf{60.0}\,/\,5.6\,/\,34.4 \\

\hspace{1.5em} \name\ vs.\ Vision 
& \textbf{84.3}\,/\,8.6\,/\,7.1
& \textbf{91.4}\,/\,7.1\,/\,1.4
& \textbf{92.9}\,/\,2.9\,/\,4.3 \\

\midrule

\rowcolor{gray!10}\multicolumn{4}{l}{\textit{Claude-haiku-4.5}} \\
\midrule
\hspace{1.5em} \name\ vs.\ Raw    
& \textbf{84.4}\,/\,7.8\,/\,7.8
& \textbf{63.3}\,/\,22.2\,/\,14.4
& \textbf{80.0}\,/\,12.2\,/\,7.8 \\

\hspace{1.5em} \name\ vs.\ Vision 
& \textbf{91.4}\,/\,5.7\,/\,2.9
& \textbf{97.1}\,/\,1.4\,/\,1.4
& \textbf{92.9}\,/\,5.7\,/\,1.4 \\

\midrule

\rowcolor{gray!10}\multicolumn{4}{l}{\textit{Qwen3.5-9b}} \\
\midrule
\hspace{1.5em} \name\ vs.\ Raw    
& \textbf{79.2}\,/\,8.3\,/\,12.5
& \textbf{73.6}\,/\,12.5\,/\,13.9
& \textbf{79.2}\,/\,11.1\,/\,9.7 \\

\hspace{1.5em} \name\ vs.\ Vision 
& \textbf{88.6}\,/\,2.9\,/\,8.6
& \textbf{100.0}\,/\,0.0\,/\,0.0
& \textbf{88.6}\,/\,11.4\,/\,0.0 \\

\bottomrule
\end{tabular}

\caption{\textbf{Full results for human evaluation on ETTh1 dataset.} Full Win/Tie/Lose ratios corresponding to the summary results in Table~\ref{tab:pairwise_comparison}. Each row compares \name\ against a baseline; \textbf{Win}/Tie/Lose values are reported from the perspective of \name. Fid.\ and Pre.\ denote fidelity and preservation, respectively.}

\label{tab:pairwise_comparison_full}
\end{table*}
\define{Annotator Recruitment} We recruited a total of 10 annotators with prior knowledge in time-series analysis. The annotators were evenly split into two groups: 5 evaluated \name against the raw string representation, and the remaining 5 evaluated \name against the vision-based representation. We paid \$6 USD per annotator for their participation, which took approximately 40 minutes to complete.

\begin{figure}[!h]
    \centering
    \includegraphics[width=0.85\columnwidth]{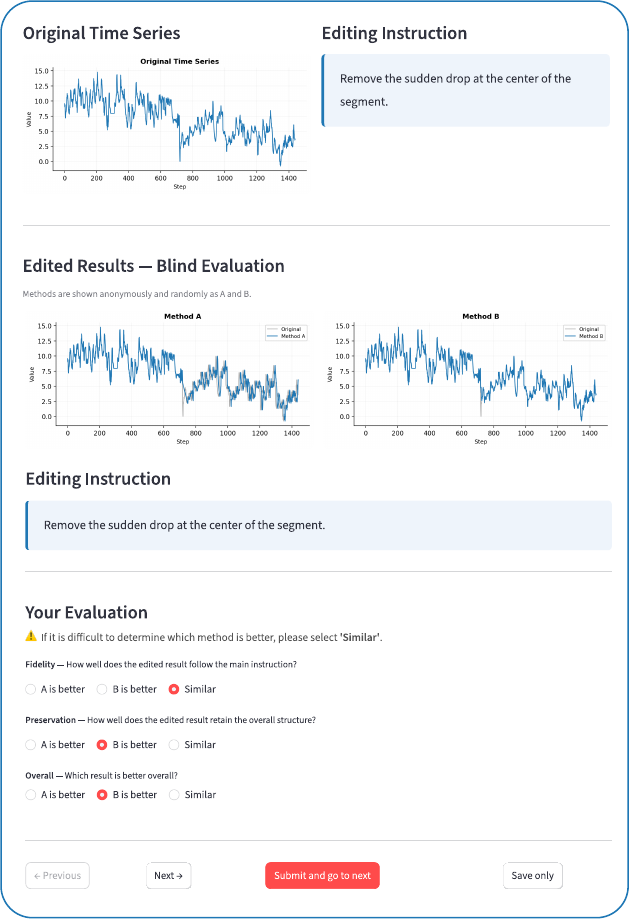}
    \caption{\define{Streamlit-based Annotation Framework} We capture the annotation interface used in our experiment.}
    \label{fig:Human_evaluation}
\end{figure}

\define{Evaluation Method} We built a streamlit-based annotation framework, where for each evaluation sample, annotators are shown three pieces of information: the original time series, a natural-language editing instruction, and two edited time series produced by two different methods. As illustrated in~\cref{fig:Human_evaluation}, the original time series is provided together with the instruction describing how the series should be modified. The two candidate edited results are then displayed side by side as Method A and Method B. To ensure a fair comparison, the evaluation is conducted in a blind and randomized manner. Specifically, one of the two edited results is generated by \name, while the other is generated by a baseline method, either the raw time-series or the vision-based baseline. The method identities are not revealed to the annotators. In addition, the order of the two results is randomly shuffled for each sample.

\define{Evaluation} Each sample is evaluated along three criteria: Fidelity, Preservation, and Overall Preference. Fidelity measures how accurately the edited time series follows the requested modification in the instruction. In other words, annotators are asked to judge which result better reflects the intended edit, such as increasing, decreasing, shifting, or otherwise modifying the specified region or pattern. Preservation measures whether the parts of the time series that are not mentioned in the instruction remain unchanged. This criterion is important because a desirable editing method should modify only the instructed aspects while preserving the remaining temporal structure of the original series. Finally, Overall Preference asks annotators to choose the result that is better overall, considering both instruction fidelity and preservation of irrelevant regions. The Similar option is provided when the two edited results are difficult to distinguish or when neither result is clearly better under the corresponding criterion.

\define{Results} We aggregate the annotations by computing the win rate of \name against each baseline. A win is counted when the annotator selects the result produced by \name over the competing baseline for a given criterion. Responses marked as \texttt{Similar} are treated as ties and are not counted as wins for either method. The resulting win rates are reported in~\cref{tab:pairwise_comparison_full}. A win rate above 50\% indicates that human annotators more frequently prefer \name over the corresponding baseline. The results show that \name is preferred over both baselines, with a particularly large margin against the vision-based baseline. This indicates that vision-based representations may struggle to support such fine-grained instructions, whereas \name more effectively performs targeted edits while preserving the remaining structure of the original signal.

\begin{figure*}[t]
    \centering
    \includegraphics[width=0.8\textwidth]{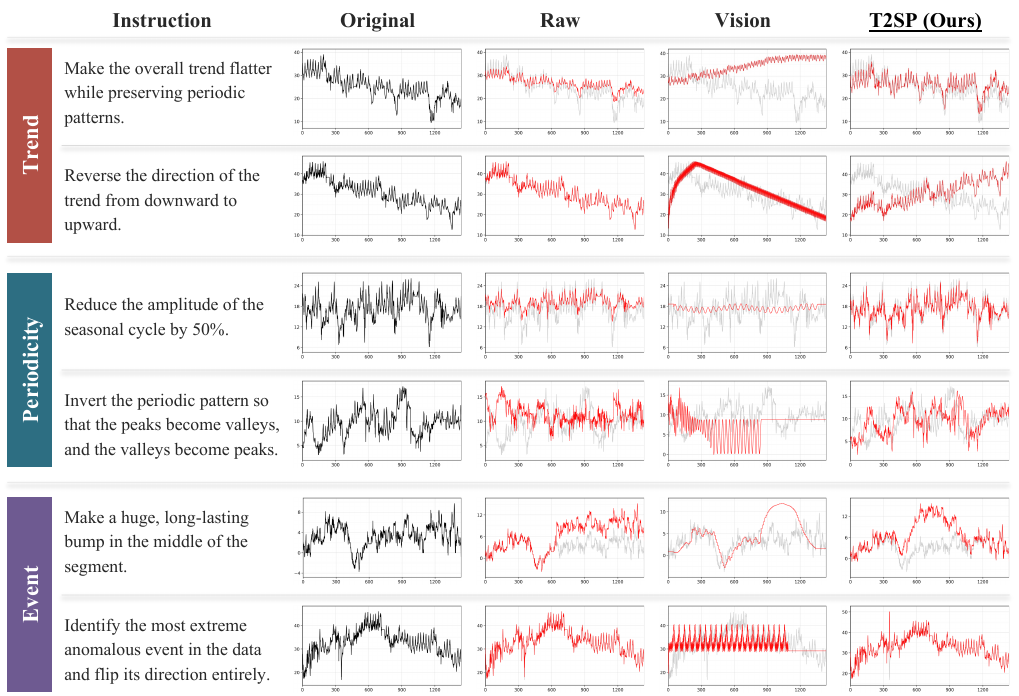}
    \caption{\define{Editing results on ETTh1 task} The grey and red lines show the original and edited time series, respectively. \name produces precise, component-level edits across trend, periodicity, and event instructions, while Raw and Vision baselines often fail to localize changes or even corrupt unrelated structure.}
    \label{fig:ETTh1_Editing}
\end{figure*}
\begin{table*}[!ht]
\centering
\small
\setlength{\tabcolsep}{7pt}
\renewcommand{\arraystretch}{1.1}
\resizebox{0.85\textwidth}{!}{%
\begin{tabular}{c|c|cccc|cccc}
\toprule
\multirow{3}{*}{N-shot} & \multirow{3}{*}{Method}
& \multicolumn{4}{c|}{\textbf{Gemini-3.1-flash-lite}}
& \multicolumn{4}{c}{\textbf{Claude-haiku-4.5}} \\
\cmidrule(lr){3-6} \cmidrule(lr){7-10}
& & \multicolumn{2}{c}{\textbf{Wafer}} & \multicolumn{2}{c|}{\textbf{ECG200}}
& \multicolumn{2}{c}{\textbf{Wafer}} & \multicolumn{2}{c}{\textbf{ECG200}} \\
& & Acc $\red\uparrow$ & F1 $\red\uparrow$
& Acc $\red\uparrow$ & F1 $\red\uparrow$
& Acc $\red\uparrow$ & F1 $\red\uparrow$
& Acc $\red\uparrow$ & F1 $\red\uparrow$ \\
\midrule
\multirow{3}{*}{\textit{zero-shot}}
& Raw
& 0.465 & 0.360
& 0.560 & 0.476
& \textbf{0.585} & \textbf{0.532}
& 0.540 & 0.452 \\
& Vision
& 0.495 & 0.363
& 0.570 & 0.402
& 0.540 & 0.467
& \textbf{0.590} & 0.485 \\
& \cellcolor{imprcolor}\name (Ours)
& \cellcolor{imprcolor}\textbf{0.505} & \cellcolor{imprcolor}\textbf{0.368}
& \cellcolor{imprcolor}\textbf{0.660} & \cellcolor{imprcolor}\textbf{0.587}
& \cellcolor{imprcolor}0.480 & \cellcolor{imprcolor}0.386
& \cellcolor{imprcolor}0.550 & \cellcolor{imprcolor}\textbf{0.529} \\
\cmidrule(lr){1-10}
\multirow{3}{*}{\textit{1-shot}}
& Raw
& 0.563 & 0.544
& 0.640 & 0.506
& \textbf{0.555} & \textbf{0.553}
& 0.630 & 0.432 \\
& Vision
& 0.415 & 0.405
& 0.630 & 0.432
& 0.515 & 0.445
& 0.630 & 0.563 \\
& \cellcolor{imprcolor}\name (Ours)
& \cellcolor{imprcolor}\textbf{0.620} & \cellcolor{imprcolor}\textbf{0.611}
& \cellcolor{imprcolor}\textbf{0.670} & \cellcolor{imprcolor}\textbf{0.526}
& \cellcolor{imprcolor}0.550 & \cellcolor{imprcolor}0.549
& \cellcolor{imprcolor}\textbf{0.710} & \cellcolor{imprcolor}\textbf{0.703} \\
\cmidrule(lr){1-10}
\multirow{3}{*}{\textit{3-shot}}
& Raw
& 0.630 & 0.617
& 0.620 & 0.382
& 0.590 & 0.571
& 0.580 & 0.554 \\
& Vision
& 0.495 & 0.412
& 0.670 & 0.474
& 0.650 & 0.648
& 0.590 & 0.580 \\
& \cellcolor{imprcolor}\name (Ours)
& \cellcolor{imprcolor}\textbf{0.835} & \cellcolor{imprcolor}\textbf{0.835}
& \cellcolor{imprcolor}\textbf{0.690} & \cellcolor{imprcolor}\textbf{0.634}
& \cellcolor{imprcolor}\textbf{0.825} & \cellcolor{imprcolor}\textbf{0.825}
& \cellcolor{imprcolor}\textbf{0.650} & \cellcolor{imprcolor}\textbf{0.645} \\
\cmidrule(lr){1-10}
\multirow{3}{*}{\textit{5-shot}}
& Raw
& 0.690 & 0.689
& 0.630 & 0.469
& 0.770 & 0.770
& 0.470 & 0.467 \\
& Vision
& 0.520 & 0.417
& 0.590 & 0.516
& 0.690 & 0.689
& 0.550 & 0.549 \\
& \cellcolor{imprcolor}\name (Ours)
& \cellcolor{imprcolor}\textbf{0.850} & \cellcolor{imprcolor}\textbf{0.849}
& \cellcolor{imprcolor}\textbf{0.710} & \cellcolor{imprcolor}\textbf{0.596}
& \cellcolor{imprcolor}\textbf{0.850} & \cellcolor{imprcolor}\textbf{0.849}
& \cellcolor{imprcolor}\textbf{0.740} & \cellcolor{imprcolor}\textbf{0.721} \\
\cmidrule(lr){1-10}
\multirow{3}{*}{\textit{20-shot}}
& Raw
& 0.650 & 0.633
& 0.650 & 0.442
& 0.845 & 0.843
& 0.610 & 0.588 \\
& Vision
& 0.560 & 0.546
& 0.650 & 0.442
& 0.715 & 0.715
& 0.700 & 0.665 \\
& \cellcolor{imprcolor}\name (Ours)
& \cellcolor{imprcolor}\textbf{0.895} & \cellcolor{imprcolor}\textbf{0.894}
& \cellcolor{imprcolor}\textbf{0.730} & \cellcolor{imprcolor}\textbf{0.668}
& \cellcolor{imprcolor}\textbf{0.900} & \cellcolor{imprcolor}\textbf{0.899}
& \cellcolor{imprcolor}\textbf{0.760} & \cellcolor{imprcolor}\textbf{0.736} \\
\bottomrule
\end{tabular}%
}
\caption{\define{Caption-based classification across LLM backbones}
Comparison of Gemini-3.1-flash-lite and Claude-haiku-4.5 under the same datasets, few-shot protocol, and sampled test sets. For each representation, captions are first generated from the time series, and classification is performed using only the generated captions. Acc and F1 denote accuracy and macro-F1, respectively.}
\label{tab:caption_downstream_combined}
\end{table*}

\subsection{Time-series Editing Examples}
We provide examples of the edited time series from the ETTh1 dataset in~\cref{fig:ETTh1_Editing}.
\clearpage
\section{Time-series Captioning}
\label{appendix:ts_caption_full}
We provide the full captioning results in~\cref{tab:caption_downstream_combined}.

\subsection{Time-series Captioning Examples}

\begin{figure}[!h]
    \centering
    \includegraphics[width=\columnwidth]{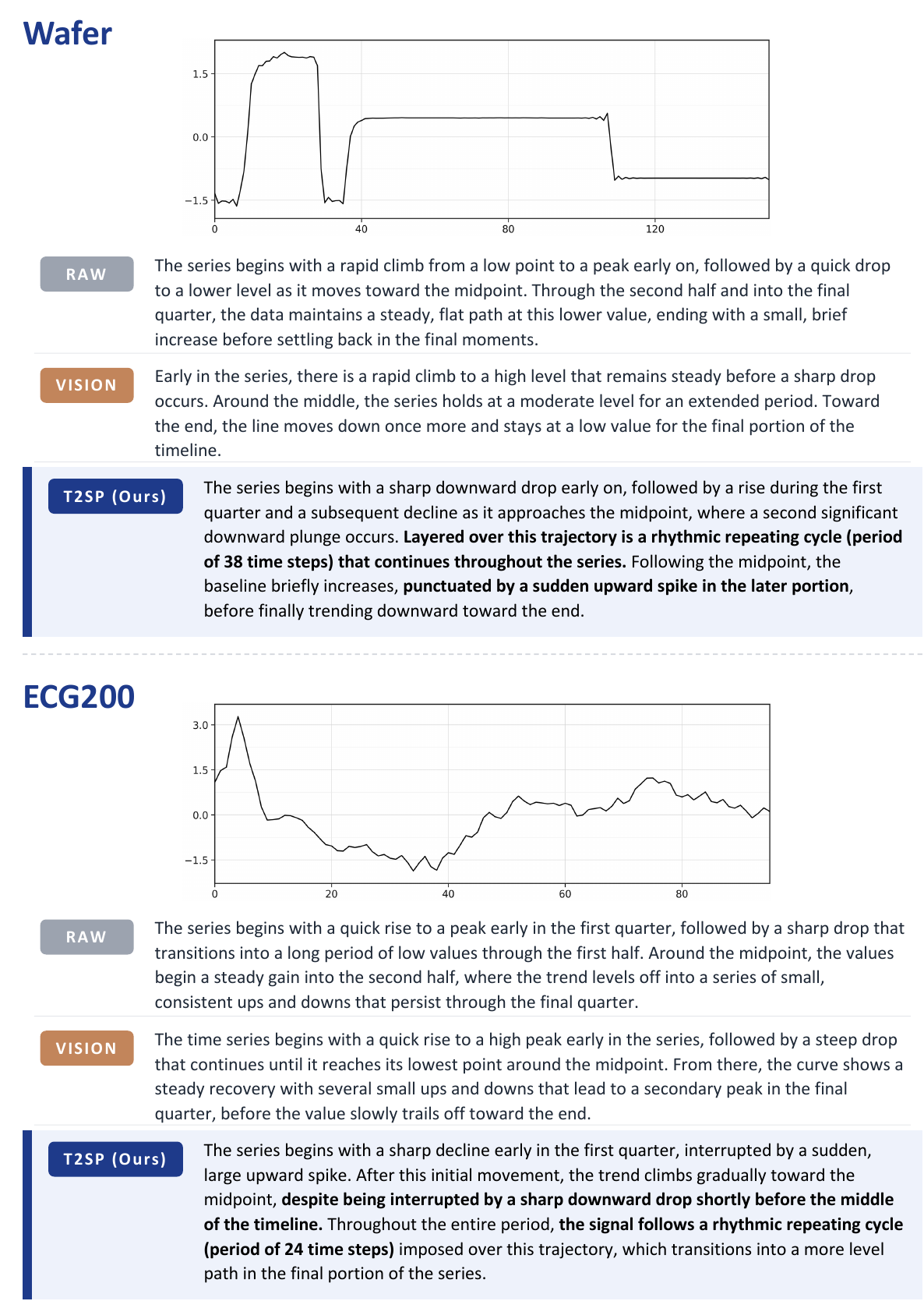}
    \caption{\define{Captioning results}}
    \label{fig:UCR_Captioning}
\end{figure}

\section{Prompts}

\begin{lstlisting}[style=promptstyle, caption={Prompt for Editing Task (TSEdit-Trend)}, label={lst:prompt}]
=========================================
Sample ID : trend_L32_0000
Category  : trend
Edit type : edit_flatten
Instruction: Flatten the trend to be constant over time.
=========================================
You are a time-series program editing expert.
Your task is to modify the symbolic program that represents a time series
so that it satisfies the editing instruction below.

# Description of the structured program representation
The structured program representation is composed of the below interpretable components:
- BSplineTrend(degree, knots, coeff)
    - represents the smooth baseline trend over time
    - degree: the degree of the B-spline (e.g., 1 for linear, 3 for cubic. Should always be >= 1)
    - knots: define where the trend is allowed to change its shape (change points in time)
    - uniform_knots: (used only when knot_mode=fixed) human-readable representative knot positions; knots is the full spline knot vector used internally, often with repeated boundary knots
    - coeff:
        - when n_knots == 2: represent the endpoint values of a global linear trend, i.e., coeff = [at_{zero} + b, at_{one} + b] for y = at + b
        - when n_knots > 2: define the spline as a linear combination y(t) = sum_i c_i B_i(t), where coeff = c_i are the basis weights controlling local shape
    - knot_ts_value: the time-series value at each knot. This is the actual value (containing the noise).
- Sinusoid(amplitude, period, phase[, start_time, end_time])
    - represents periodic oscillations
    - amplitude: strength of oscillation
    - period: length of one cycle
    - phase: horizontal shift of the wave
    - r2: goodness-of-fit of the sinusoidal component (higher means the periodic pattern explains the signal well)
    - optional start_time/end_time indicate the knot (time) interval where the oscillation is active
- GaussianEvent(center, width, amplitude)
    - represents localized transient events
    - center: where the event occurs
    - width: how spread the event is over time
    - amplitude: magnitude of the event
- SpikeEvent(index, amplitude)
    - represents instantaneous sharp anomalies
    - index: time of occurrence
    - amplitude: magnitude of the spike
- Residual(std)
    - represents unexplained noise
    - std: noise intensity
The final time series is obtained by combining all components additively.
Each component has a clear semantic meaning, and the structured program representation provides a structured, interpretable representation of the signal.


# Editing Instruction:
Flatten the trend to be constant over time.

# Original Program:
Series(
    trend=BSplineTrend(
        degree=1,
        smoothness=5,
        knot_mode='fixed',
        knots=[0, 0, 31, 31],
        coeff=[-0.1972, 1.4],
        n_knots=2,
        uniform_knots=[0, 31],
        knot_ts_values=[-0.3627, 1.076]
    ),
    periodic=[
        Sinusoid(amplitude=1.7361, period=4, phase=0.6494, frequency=0.25, r2=0.9812)
    ],
    events=[
        # none
    ],
    noise=Residual(std=0.1698, mean=0)
)


# Rules:

- Provide your reasoning in the <Reason> ... </Reason> section, but do NOT include any code there.
- Wrap the program in <program> ... </program> tags.

Your answer should be formatted as follows:
<Reason>
Provide your reasoning here, explaining how you modified the original program to satisfy the editing instruction.
Be specific about which parts of the program you changed and why.
</Reason>
<program>
Output your edited program code here. Return ONLY the modified program code.
</program>
\end{lstlisting}

\section{Use of LLM Assistance} 
We used Claude for sentence-level editing and for coding assistance. All scientific content, experimental design, and conclusions are the authors' own.
\newpage

\end{document}